\documentclass{article}
\usepackage{spconf,amsmath,graphicx}
\usepackage{enumitem} 
\usepackage{amssymb}
\usepackage{booktabs}
\usepackage{multirow}
\usepackage{graphicx}
\usepackage{svg}

\setlist{topsep=0pt, partopsep=0pt, parsep=0pt}
\usepackage{url} 

\title{LFA: Layer Feature Attention for Run-time Introspection of \\2D Object Detectors in Automated Driving}

\name{Mert Keser\textsuperscript{\textdagger\textdaggerdbl}, Alois Knoll\textsuperscript{\textdaggerdbl}\thanks{The research leading to these results is funded by the German Federal Ministry for Economic Affairs and Energy within the project ``NXT GEN AI METHODS -- Generative Methoden f\"ur Perzeption, Pr\"adiktion und Planung''. The authors would like to thank the consortium for the successful cooperation.}}
\address{AUMOVIO SE, Germany\textsuperscript{\textdagger} \\ Technical University of Munich, Germany\textsuperscript{\textdaggerdbl}}

\address{Aumovio SE, Germany\textsuperscript{\textdagger}\\
	Technical University of Munich, Germany\textsuperscript{\textdagger}\textsuperscript{\textdaggerdbl}}

\begin{document}
%
\maketitle

\begin{abstract}

Reliable object detection is critical for automated driving, yet even state-of-the-art detectors inevitably make errors that can compromise safety. Introspection methods that predict detector failures enable safer deployment by triggering fallback mechanisms or alerting human operators. However, existing approaches rely solely on last-layer features or hand-crafted statistics, discarding valuable information from earlier layers that capture different levels of visual abstraction. We propose Layer Feature Attention (LFA), a lightweight introspection method that learns to aggregate features from multiple backbone layers through an attention mechanism. Our key insight is that detection errors manifest differently across feature hierarchies—low-level layers capture fine-grained details essential for detecting small or occluded objects, while high-level layers encode semantic information for scene understanding. LFA learns layer importance weights end-to-end, enabling both improved error prediction and interpretable analysis of which feature levels are most indicative of detector failures. Extensive experiments on KITTI and BDD100K demonstrate that LFA achieves state-of-the-art introspection performance, outperforming single-layer baselines across multiple detector architectures.

\end{abstract}

\vspace{-0.8em}
\section{Introduction}
\label{sec:intro}

Accurate perception of the surrounding environment is of paramount importance for the safe operation of automated driving (AD) systems \cite{yurtsever2020survey}. Within the perception stack, object detection provides instance-level information by identifying and localizing traffic participants such as vehicles, pedestrians, and cyclists. Despite substantial progress on benchmark datasets \cite{geiger2012we, yu2020bdd100k}, deep neural network (DNN)-based detectors remain vulnerable in real-world deployment. In practice, distribution shifts between training and operational domains, adverse environmental conditions, and rare or long-tail scenarios frequently lead to detection failures \cite{keser2025benchmarking, breitenstein2023does}. Since perfect detection performance cannot be guaranteed, recent research has increasingly emphasized introspection—the ability of a perception system to assess the reliability of its own outputs at run time and to indicate potential failure cases \cite{rahman2021run}. This need is further reinforced by recent regulatory developments, including the EU AI Act \cite{EU_AI_Act_2024} and ISO/PAS 8800:2024 \cite{ISO_PAS_8800_2024}, which explicitly call for run-time monitoring mechanisms in safety-critical AI systems. 

Introspection for object detection aims to predict, at run-time, whether the detector has produced erroneous outputs for a given input frame, thereby enabling the deployment of appropriate fallback strategies such as human takeover requests or minimum risk maneuvers \cite{yatbaz2023introspection}. In principle, an introspection mechanism can leverage information from any stage of the detection pipeline — the raw input, intermediate feature representations, or the final predictions \cite{rahman2021run}. The majority of prior work operates at the output level, using softmax confidence scores or hand-crafted statistics derived from detector outputs as proxies for prediction reliability. However, such output-level cues are often poorly calibrated and can provide misleading estimates of actual model uncertainty \cite{miller2018dropout}. Motivated by these limitations, recent work has increasingly explored feature-based introspection methods that train classifiers on the detector's internal backbone activations \cite{yatbaz2024run, rahman2021per}. Nevertheless, existing feature-based approaches typically rely on representations extracted from a single, manually selected backbone layer—most commonly the last—without a principled justification for this choice, thereby neglecting complementary information encoded at other levels of the feature hierarchy.


\setlength{\abovedisplayskip}{3pt}
\setlength{\belowdisplayskip}{3pt}
\setlength{\abovedisplayshortskip}{3pt}
\setlength{\belowdisplayshortskip}{3pt}

\setlength{\textfloatsep}{6pt plus 1pt minus 2pt}
\setlength{\floatsep}{6pt plus 1pt minus 2pt}
\setlength{\intextsep}{6pt plus 1pt minus 2pt}
   
\begin{figure*}[t]
  \centering
  \includegraphics[width=0.72\textwidth]{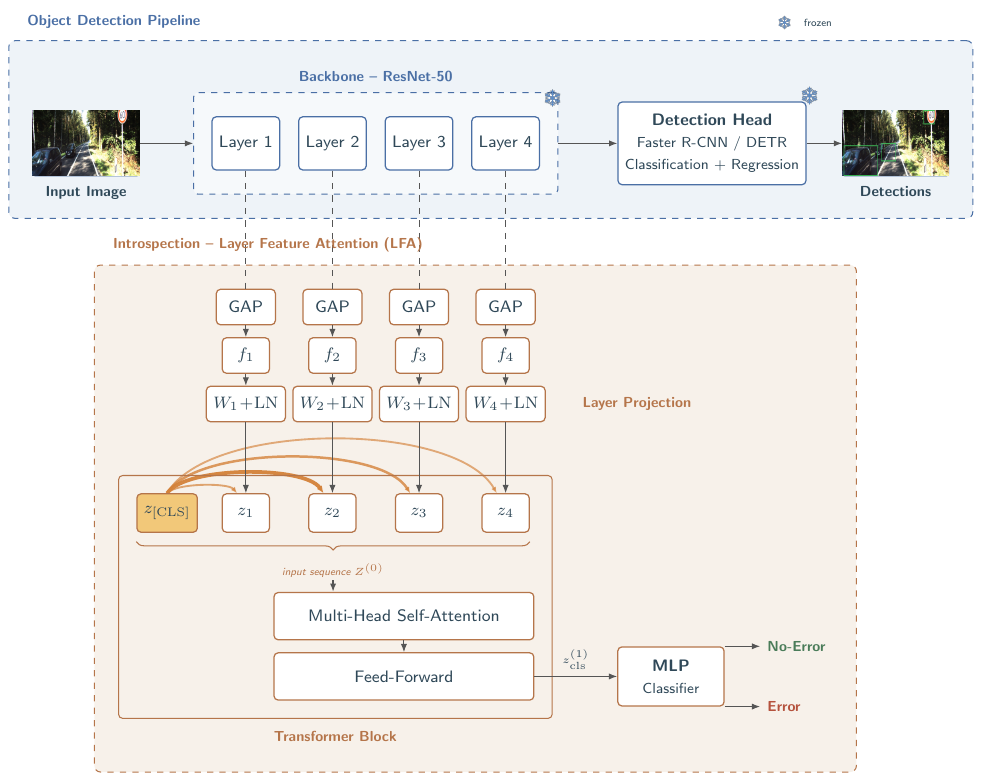}
  \caption{Overview of the LFA framework. The frozen detector (top) extracts multi-layer features from a ResNet-50 backbone; LFA (bottom) applies GAP and layer-specific projections, then aggregates the embeddings via a transformer block to predict frame-level detection errors.}
  \label{fig:architecture}
\end{figure*}

In this paper, we propose Layer Feature Attention (LFA), an introspection method that moves beyond the single-layer paradigm by learning to aggregate features from multiple backbone layers through an attention mechanism. LFA treats each backbone layer's feature representation as an individual token and employs a lightweight transformer module to learn the relevance of each layer for the error prediction task. This formulation is grounded in the observation that different backbone layers encode fundamentally different levels of visual abstraction, and that detection errors do not manifest uniformly across this hierarchy. By learning to selectively attend to the most informative layers, LFA adapts its feature aggregation strategy to the characteristics of each dataset — an ability that single-layer methods inherently lack. Beyond improved accuracy, this design also offers interpretability: the learned attention weights directly reveal which levels of the feature hierarchy are most indicative of detector failures.

Our contributions are threefold:
\begin{enumerate}[leftmargin=1.5em, label=(\roman*), itemsep=0pt, parsep=0pt, topsep=0pt]
\item We propose Layer Feature Attention (LFA), an introspection framework that aggregates multi-layer backbone features via a learned attention mechanism, moving beyond the single-layer paradigm adopted by prior approaches.
\item We provide an interpretable analysis of the learned attention weights, demonstrating that the relative importance of feature hierarchy levels is dataset-dependent and varies across driving scenarios.
\item We extensively evaluate LFA on established autonomous driving benchmarks and across different object detector architectures, achieving state-of-the-art introspection performance and outperforming existing single-layer baselines.
\end{enumerate}

\vspace{-0.8em}
\section{Related Work}
\label{sec:related}

We review introspection methods for object detection (Sec.~\ref{sec:rw_introspection}) and situate our approach within feature-based introspection methods (Sec.~\ref{sec:rw_feature}).

\subsection{Introspection for Object Detection}
\label{sec:rw_introspection}
Introspection methods for object detection in AD
can be broadly categorized according to the type of information
they exploit. Confidence-based approaches leverage softmax scores
or Bayesian uncertainty estimates to flag unreliable predictions
\cite{harakeh2020bayesod, miller2018dropout}. However, such signals
are often poorly calibrated and may provide misleading estimates
of prediction reliability \cite{miller2018dropout}. Metric-based
methods instead aim to predict frame-level performance indicators,
such as mAP, using auxiliary networks
\cite{rahman2021run, rahman2021per}. Inconsistency-based approaches
identify failures by detecting disagreements across multiple sensors
or perception algorithms \cite{ramanagopal2018failing}, while
concept-based methods verify predictions against semantically
meaningful concepts \cite{keser2023interpretable}. While effective
in specific settings, these categories either rely on unreliable
output-level cues or require access to additional modalities or
supervision. In contrast, a more recent and promising direction is
feature-based introspection, which directly exploits the detector's
internal representations to assess its own reliability.

\subsection{Feature-based Introspection}
\label{sec:rw_feature}
Yatbaz et al.~\cite{yatbaz2024run} proposed Learned Feature
Representations (LFR), which extract activations from the final
backbone layer and train a classifier directly on these features,
bypassing the need for hand-crafted feature engineering. Building
on this idea, LF-ASH \cite{yatbaz2023introspection} applies Activation
Shaping (ASH) \cite{djurisic2022extremely} to suppress less informative
activations within the extracted layer, thereby improving
discriminability. In the context of LiDAR-based 3D object detection,
a recent extension \cite{yatbaz2025multi} investigated the role of
activations from different backbone layers and proposed concatenating
early, intermediate, and final layer features for introspection.
While this represents an important step toward multi-layer
introspection, the feature aggregation strategy remains fixed and
does not account for the varying relevance of different layers.
Moreover, this approach targets 3D point cloud detectors, leaving
multi-layer introspection for 2D object detection unexplored.
LFA addresses these limitations by introducing a learned attention
mechanism that adaptively weights the contribution of each backbone
layer for the error prediction task in 2D object detection.

\section{Methodology}
\label{sec:methodology}

We introduce Layer Feature Attention (LFA) (Sec.~\ref{sec:lfa}), 
and describe the introspection framework for its training and evaluation (Sec.~\ref{sec:introspection_training}).

\subsection{Layer Feature Attention}
\label{sec:lfa}

LFA takes GAP-pooled feature vectors from all backbone layers and learns to aggregate them via a transformer attention mechanism for frame-level error prediction.

\smallskip
\noindent\textbf{Layer Projection.}
Given an object detector with backbone $B$, we extract feature
maps from $L$ intermediate layers and apply global average
pooling (GAP) to obtain compact feature vectors
$\mathbf{f}_\ell \in \mathbb{R}^{C_\ell}$ for each layer
$\ell \in \{1, \dots, L\}$, where $C_\ell$ denotes the channel
dimension. Since feature dimensionalities differ across backbone
stages (e.g., $C_\ell \in \{256, 512, 1024, 2048\}$ for a
ResNet-50 backbone~\cite{he2016deep}), we project each vector into
a shared embedding space:
\begin{equation}
    \mathbf{z}_\ell = \mathrm{LN}\!\left(\mathbf{W}_\ell
    \mathbf{f}_\ell + \mathbf{b}_\ell\right),
    \quad \mathbf{z}_\ell \in \mathbb{R}^{d},
\label{eq:projection}
\end{equation}
where $\mathbf{W}_\ell \in \mathbb{R}^{d \times C_\ell}$ and
$\mathbf{b}_\ell$ are learnable, layer-specific projection
parameters, and $\mathrm{LN}$ denotes layer normalization
\cite{ba2016layer}.

\smallskip
\noindent\textbf{Transformer Attention.}
We form a sequence by prepending a learnable classification
token $\mathbf{z}_{\texttt{cls}} \in \mathbb{R}^{d}$ to the
projected layer embeddings and add learnable layer embeddings
$\mathbf{e}_i$ to encode token identity:
\begin{equation}
    \mathbf{Z}^{(0)} =
    \left[\mathbf{z}_{\texttt{cls}},\,
    \mathbf{z}_1, \dots, \mathbf{z}_L \right]
    +
    \left[\mathbf{e}_0,\, \mathbf{e}_1, \dots, \mathbf{e}_L \right],
\label{eq:sequence}
\end{equation}
where $\mathbf{Z}^{(0)} \in \mathbb{R}^{(L+1) \times d}$. The
sequence is processed by a single pre-norm transformer block
comprising multi-head self-attention (MHSA) and a feed-forward
network (FFN):
\begin{align}
    \hat{\mathbf{Z}} &=
    \mathbf{Z}^{(0)} +
    \mathrm{MHSA}\!\left(\mathrm{LN}\!\left(\mathbf{Z}^{(0)}\right)\right),
    \label{eq:mhsa} \\
    \mathbf{Z}^{(1)} &=
    \hat{\mathbf{Z}} +
    \mathrm{FFN}\!\left(\mathrm{LN}\!\left(\hat{\mathbf{Z}}\right)\right).
    \label{eq:ffn}
\end{align}

\smallskip
\noindent\textbf{Classification.}
The output corresponding to the classification token,
$\mathbf{z}^{(1)}_{\texttt{cls}} = \mathrm{LN}\!\left(
\mathbf{Z}^{(1)}_{0}\right)$, serves as the aggregated
representation and is passed to a two-layer MLP to produce the
final error prediction:
\begin{equation}
    \hat{y} =
    \mathrm{MLP}\!\left(\mathbf{z}^{(1)}_{\texttt{cls}}\right)
    \in \mathbb{R}^{2}.
\label{eq:classifier}
\end{equation}

\smallskip
\noindent\textbf{Layer Interpretability.}
Although the attention weights are not used in the
classification itself, they provide a post-hoc measure of layer
relevance. Specifically, the attention weights from the
classification token to the layer tokens are given by
\begin{equation}
    \boldsymbol{\alpha} =
    \mathrm{softmax}\!\left(
    \frac{\mathbf{q}_{\texttt{cls}} \mathbf{K}^\top}{\sqrt{d_h}}
    \right)
    \in \mathbb{R}^{L+1},
\label{eq:attention}
\end{equation}
where $\mathbf{q}_{\texttt{cls}}$ denotes the query corresponding
to the classification token, $\mathbf{K}$ the key matrix, and
$d_h = d / H$ the per-head dimension. The resulting distribution
$[\alpha_1, \dots, \alpha_L]$ indicates the relative importance
assigned to each backbone layer for the error prediction task.

\subsection{Introspection Training}
\label{sec:introspection_training}

\noindent\textbf{Error Labeling.}
Prior introspection methods for 2D object detection~\cite{rahman2021per, yatbaz2024run} learn the relationship between backbone activation patterns and frame-level mean average precision~(mAP). However, mAP-based labeling can mask individual missed objects due to per-class averaging~\cite{yatbaz2023introspection}. Following~\cite{yatbaz2023introspection}, we adopt a false-negative~(FN) based labeling strategy that directly targets missed detections.
Given ground-truth boxes $\mathcal{G} = \{g_1, \dots, g_M\}$ and detector predictions $\mathcal{D} = \{d_1, \dots, d_N\}$ for a frame~$I$, we perform class-aware greedy matching: each detection is paired with the ground-truth box of the same class yielding the highest intersection-over-union~(IoU), subject to $\tau_{\mathrm{IoU}} = 0.5$. Any ground-truth object that remains unmatched is a false negative, and the binary frame-level label is defined as
\begin{equation}
    y =
    \begin{cases}
        1 \;\; (\textit{error}), & \text{if } \exists\, g \in \mathcal{G} \text{ unmatched},\\[2pt]
        0 \;\; (\textit{no\text{-}error}), & \text{otherwise}.
    \end{cases}
\label{eq:fn_label}
\end{equation}

\smallskip
\noindent\textbf{Introspection Pipeline.}

Our introspection pipeline comprises three stages. First, the object detector is fine-tuned from COCO-pretrained weights on the target driving dataset and then frozen. Second, the frozen detector generates predictions on a held-out validation split, which are compared against ground-truth using the FN-based labeling above to produce binary error labels; the resulting pairs of GAP-pooled features (Sec.~\ref{sec:lfa}) and labels train the introspection model with a weighted cross-entropy loss (error-class weight $w = n_{\text{neg}} / n_{\text{pos}}$ to address class imbalance). Third, the trained model is evaluated on the test split, unseen during both detector fine-tuning and introspection training.

\section{Experiments}
\label{sec:experiments}

\subsection{Experimental Setup}
\label{sec:setup}

\noindent\textbf{Datasets.}
We evaluate our approach on two autonomous driving benchmarks.
\textit{KITTI}~\cite{geiger2012we} provides 7,481 labeled urban
driving images with 2D bounding box annotations; since
the official test set labels are not publicly available,
we follow~\cite{yatbaz2024run} and partition the labeled set into
60\%/20\%/20\% splits for training, validation, and testing.
\textit{BDD100K}~\cite{yu2020bdd100k} provides 100K driving images
with official training, validation, and test partitions. To ensure
consistency across datasets, we merge the original object categories
into two classes: \textit{Vehicle} (car, van, truck, bus) and
\textit{People} (pedestrian, cyclist, rider),
following~\cite{yatbaz2024run}.

\smallskip
\noindent\textbf{Object Detectors.}
We evaluate with two architectures representing different detection paradigms: Faster R-CNN~\cite{ren2015faster}, a two-stage anchor-based detector\footnote{\url{https://pytorch.org/vision/stable/models/generated/torchvision.models.detection.fasterrcnn_resnet50_fpn_v2.html}}, and DETR~\cite{carion2020end}, a transformer-based end-to-end detector\footnote{\url{https://github.com/facebookresearch/detr}}. Both detectors employ a ResNet-50~\cite{he2016deep} backbone and are initialized from COCO-pretrained weights before fine-tuning on the training split of the respective driving dataset.

\smallskip
\noindent\textbf{Baselines.}
We compare LFA against three established introspection baselines:
(i)~\textit{SF} (Softmax Features)~\cite{rahman2021per}, which relies
solely on the detector’s output confidence scores as features;
(ii)~\textit{LFR} (Last Feature Raw)~\cite{yatbaz2023introspection}, which applies
global average pooling (GAP) to the final backbone layer without
additional preprocessing; and (iii)~\textit{LF-ASH} (Last Feature with
Activation Shaping)~\cite{yatbaz2024run}, which applies activation
shaping~\cite{djurisic2022extremely} to the final backbone layer prior
to GAP. For LF-ASH, we adopt pruning percentiles of 90\% on KITTI and
75\% on BDD100K, following the best-performing configurations reported
in~\cite{yatbaz2024run}. All baselines employ the same training configuration to ensure a fair comparison.

\smallskip
\noindent\textbf{Evaluation Metrics.}
We report three complementary metrics. The area under the receiver operating characteristic curve (\textit{AUROC}) measures ranking performance across all classification thresholds. \textit{F1} captures the harmonic mean of precision and recall. The false negative rate (\textit{FNR}) quantifies the proportion of true error frames that the introspection model fails to detect; a low FNR is essential for safety-critical deployment, as missed errors may lead to hazardous driving decisions.

\subsection{Results}
\label{sec:results}

\noindent\textbf{Introspection Performance.}
Tables~\ref{tab:kitti_results} and~\ref{tab:bdd_results} summarize the introspection performance on KITTI and BDD100K, respectively.

\begin{table}[t]
\centering
\caption{Introspection performance on KITTI. Best results per detector are in \textbf{bold}.}
\label{tab:kitti_results}
\setlength{\tabcolsep}{4pt}
\begin{tabular}{ll ccc}
\toprule
Detector & Method & AUROC $\uparrow$ & F1 $\uparrow$ & FNR $\downarrow$ \\
\midrule
\multirow{4}{*}{DETR}
 & SF            & 0.8544 & 0.8856 & 0.1814 \\
 & LFR           & 0.8908 & \textbf{0.9517} & \textbf{0.0595} \\
 & LF-ASH        & 0.8866 & 0.9422 & \textbf{0.0595} \\
 & LFA (Ours)    & \textbf{0.9118} & 0.8762 & 0.2100 \\
\midrule
\multirow{4}{*}{\shortstack{Faster\\R-CNN}}
 & SF            & 0.7350 & 0.6226 & 0.3125 \\
 & LFR           & 0.7932 & 0.6478 & \textbf{0.1989} \\
 & LF-ASH        & 0.7724 & 0.5845 & 0.4564 \\
 & LFA (Ours)    & \textbf{0.8422} & \textbf{0.6998} & \textbf{0.1989} \\
\bottomrule
\end{tabular}
\end{table}

\begin{table}[t]
\centering
\caption{Introspection performance on BDD100K. Best results per detector are in \textbf{bold}.}
\label{tab:bdd_results}
\setlength{\tabcolsep}{4pt}
\begin{tabular}{ll ccc}
\toprule
Detector & Method & AUROC $\uparrow$ & F1 $\uparrow$ & FNR $\downarrow$ \\
\midrule
\multirow{4}{*}{DETR}
 & SF            & 0.7979 & 0.7201 & 0.4191 \\
 & LFR           & 0.7744 & 0.7750 & 0.3332 \\
 & LF-ASH        & 0.7699 & 0.7386 & 0.3875 \\
 & LFA (Ours)    & \textbf{0.8045} & \textbf{0.8781} & \textbf{0.1322} \\
\midrule
\multirow{4}{*}{\shortstack{Faster\\R-CNN}}
 & SF            & 0.7006 & 0.7816 & 0.1650 \\
 & LFR           & 0.6973 & 0.8023 & 0.0421 \\
 & LF-ASH        & 0.6986 & 0.8004 & 0.0682 \\
 & LFA (Ours)    & \textbf{0.7161} & \textbf{0.8052} & \textbf{0.0310} \\
\bottomrule
\end{tabular}
\end{table}

On KITTI, LFA achieves the highest AUROC with both detectors, reaching 0.9118 with DETR and 0.8422 with Faster R-CNN. With Faster R-CNN, this represents gains of 4.9 and 7.0 percentage points over LFR and LF-ASH, respectively, while also yielding the best F1 (0.6998) and matching the lowest FNR (0.1989). These results indicate that aggregating features from multiple backbone layers produces a more discriminative error signal than relying on a single layer.

With DETR on KITTI, LFR and LF-ASH achieve higher F1 and lower FNR at the selected operating threshold, whereas LFA maintains a clear lead in AUROC. We attribute this behavior to the strong per-frame detection accuracy of the fine-tuned DETR, which results in fewer error frames and a highly imbalanced label distribution. Under such conditions, single-layer methods can achieve favorable threshold-dependent metrics; however, the superior AUROC of LFA demonstrates better separation between error and no-error frames across all operating points.

On BDD100K, LFA again attains the highest AUROC with both detectors (0.8045 with DETR, 0.7161 with Faster R-CNN) and achieves the best performance across all three metrics. The consistent improvements on this larger and more diverse dataset confirm that the proposed multi-layer attention mechanism generalizes beyond KITTI. Across all four detector–dataset combinations, LFA ranks first in AUROC, demonstrating the robustness of multi-layer feature aggregation across different detection architectures and data distributions.

\smallskip
\noindent\textbf{Ablation Study.}
Table~\ref{tab:ablation} isolates the contribution of learned attention in LFA. \textit{LFA\_Uniform} replaces the transformer attention with uniform averaging over the projected layer features, while \textit{LFA\_Concat} removes the attention mechanism entirely and instead concatenates all layer features before classification. Both ablations achieve similar AUROC to the full model, indicating that multi-layer information is beneficial regardless of the aggregation strategy. However, the full LFA achieves substantially lower FNR (0.1989 vs.\ 0.27+), demonstrating that learned attention is essential for minimizing missed detection errors at the operating threshold.

\begin{table}[t]
\centering
\caption{Ablation study on KITTI with Faster R-CNN.}
\label{tab:ablation}
\setlength{\tabcolsep}{5pt}
\begin{tabular}{l ccc}
\toprule
Variant & AUROC $\uparrow$ & F1 $\uparrow$ & FNR $\downarrow$ \\
\midrule
LFA (Ours)      & \textbf{0.8422} & 0.6998 & \textbf{0.1989} \\
LFA\_Uniform    & 0.8200 & 0.6997 & 0.2784 \\
LFA\_Concat     & 0.8209 & \textbf{0.7034} & 0.2746 \\
\bottomrule
\end{tabular}
\end{table}

\begin{figure}[t]
\centering
\includegraphics[width=0.85\columnwidth]{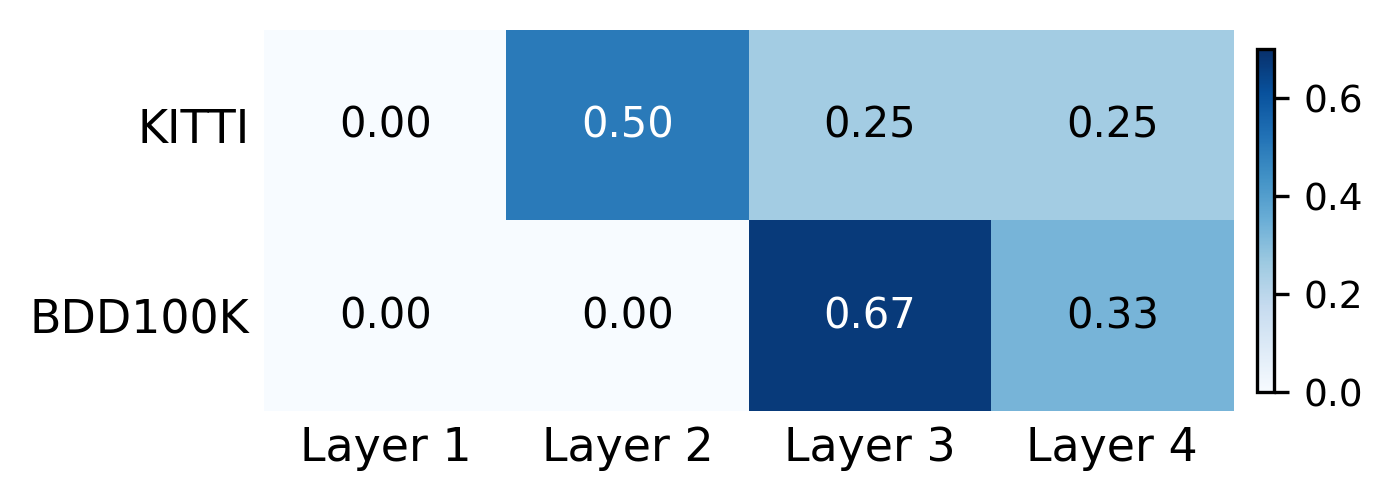}
\caption{Learned attention weights across backbone layers for
Faster R-CNN. LFA assigns dataset-dependent importance, emphasizing
mid-level features on KITTI and higher-level features on BDD100K.}
\label{fig:attention}
\end{figure}

\smallskip
\noindent\textbf{Layer Importance.}
Figure~\ref{fig:attention} visualizes the learned attention weights
across backbone layers for Faster R-CNN. On KITTI, the model assigns
the highest weight to Layer~2 (0.50), with Layers~3 and~4 sharing the
remaining attention. On BDD100K, the attention distribution shifts
toward deeper layers, with Layer~3 receiving the largest weight
(0.67). In both cases, Layer~1 receives negligible attention,
indicating that low-level features are less informative for the
error prediction task in these settings. The increased emphasis on
deeper layers for BDD100K may be attributed to the greater diversity
of recording conditions in this dataset, including variations in
weather, illumination, and scene context. Under such variability,
higher-level semantic features appear to provide more robust signals
for introspection than appearance-sensitive low-level features.

\begin{figure}[t]
\centering
\begin{tabular}{cc}
\includegraphics[width=0.48\columnwidth]{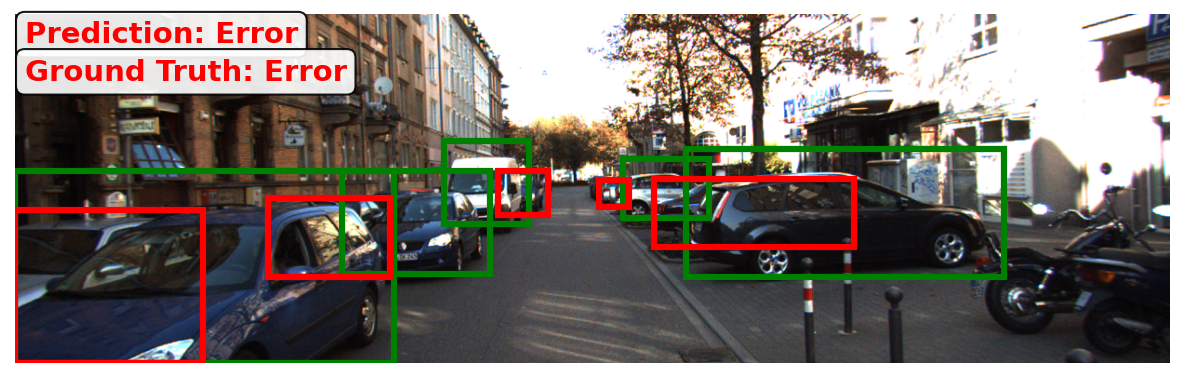} &
\includegraphics[width=0.48\columnwidth]{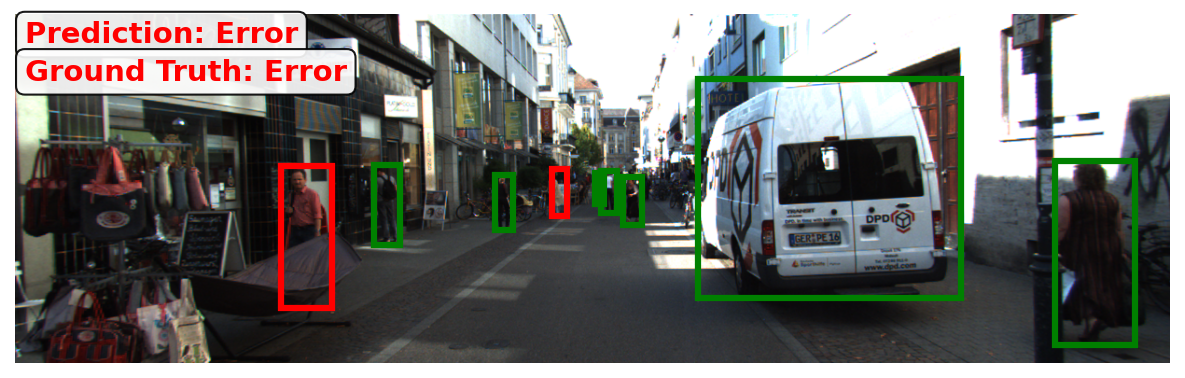} \\
{\small (a) DETR -- Error (TP)} & {\small (b) Faster R-CNN -- Error (TP)} \\[4pt]
\includegraphics[width=0.48\columnwidth]{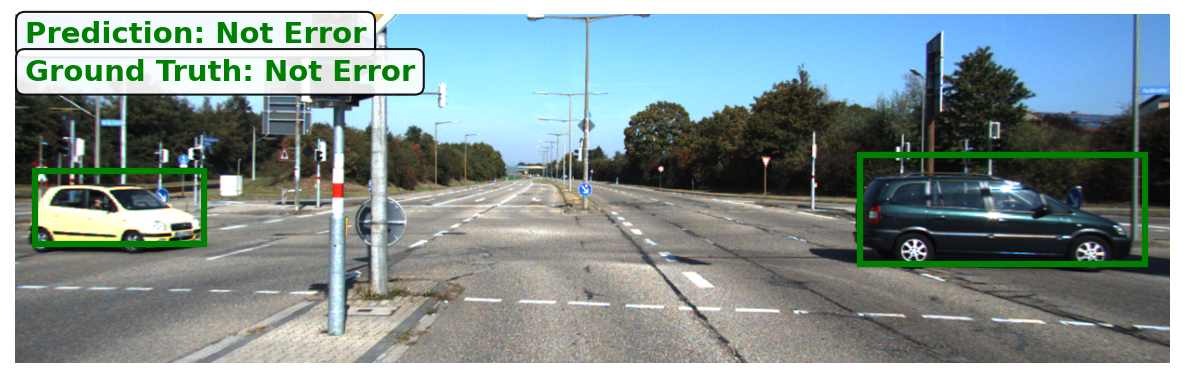} &
\includegraphics[width=0.48\columnwidth]{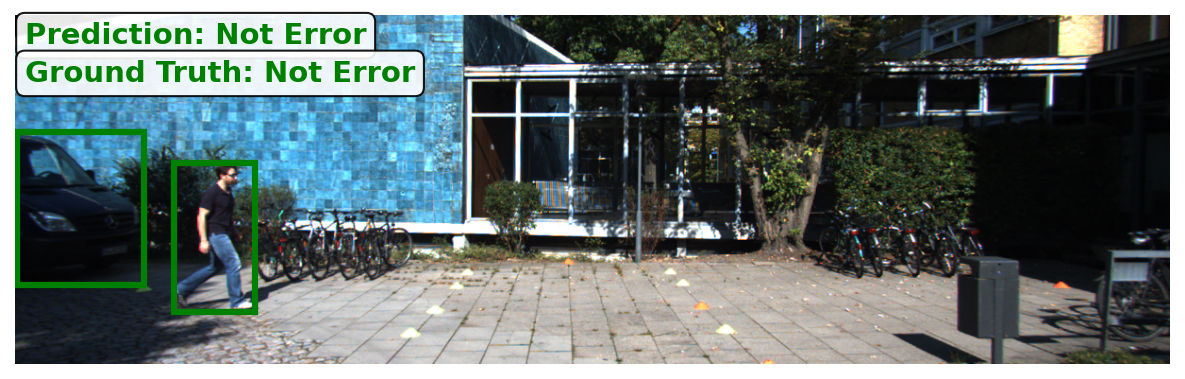} \\
{\small (c) DETR -- No Error (TN)} & {\small (d) Faster R-CNN -- No Error (TN)} \\[4pt]
\includegraphics[width=0.48\columnwidth]{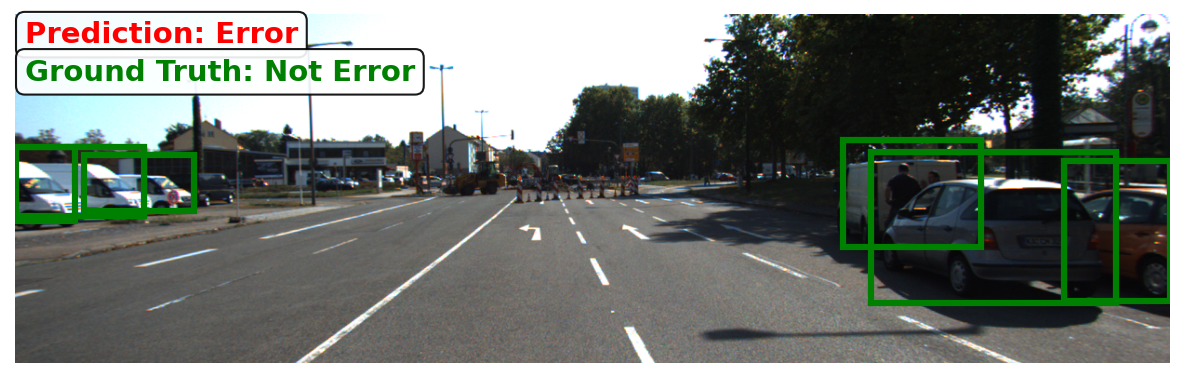} &
\includegraphics[width=0.48\columnwidth]{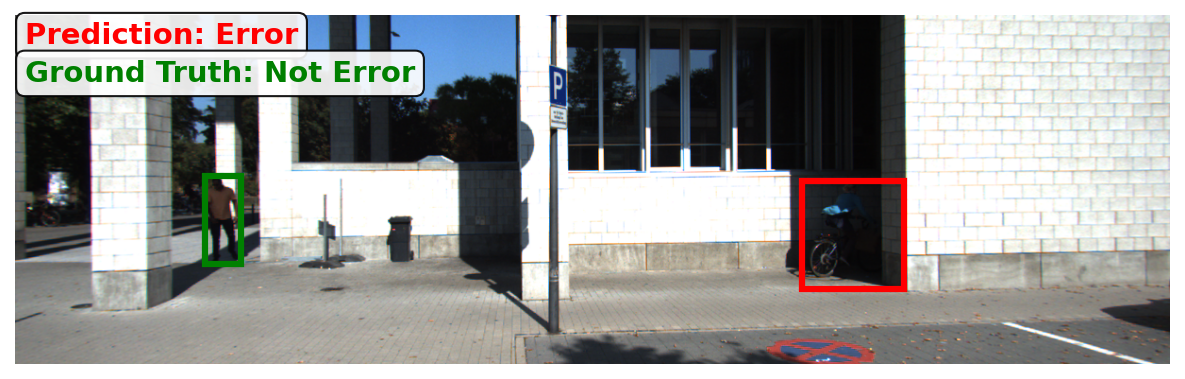} \\
{\small (e) DETR -- FP (missing GT)} & {\small (f) Faster R-CNN -- FP (missing GT)} \\
\end{tabular}
\caption{Qualitative results on KITTI. Top: error frames correctly identified (TP). Middle: no-error frames correctly identified (TN). Bottom: false positives caused by missing ground-truth annotations. Green: detections, Red: missed objects.}
\label{fig:qualitative}
\end{figure}

\smallskip
\noindent\textbf{Qualitative Examples.}
Figure~\ref{fig:qualitative} shows representative predictions from LFA on KITTI. Green boxes denote correct detections, while red boxes indicate missed objects. In the error cases (a, b), both detectors miss several vehicles and LFA correctly predicts an error. In the no-error cases (c, d), all annotated objects are detected and LFA correctly predicts no error. The bottom row (e, f) illustrates false positive cases where LFA predicts an error but the ground truth indicates no error. In (e), the detector correctly identifies distant vehicles that lack ground-truth annotations; in (f), a bicycle is detected but absent from the labels. Rather than representing true model failures, these false positives suggest that LFA can identify potential perception gaps arising from annotation incompleteness.

\smallskip
\noindent\textbf{Computational Efficiency.}
Among feature-based methods, LFR and LF-ASH employ a ResNet-18 encoder (11.2M parameters) to process spatial feature maps from the final backbone layer. In contrast, LFA applies a single-layer transformer over GAP-pooled features from all four backbone stages, requiring only 1.8M parameters—a 6$\times$ reduction. All methods operate in sub-millisecond time, adding negligible overhead to the detection pipeline \cite{yatbaz2024run}.

\section{Conclusion}
\label{sec:conclusion}

We presented Layer Feature Attention (LFA), an introspection method that aggregates features from multiple backbone layers via learned attention to predict object detection errors at the frame level. Unlike prior approaches that rely on a single layer or hand-crafted preprocessing, LFA learns to adaptively weight layer contributions, enabling the integration of complementary information across the feature hierarchy. Experiments on KITTI and BDD100K using both Faster R-CNN and DETR show that LFA achieves the highest AUROC across all settings, while yielding notable reductions in false negative rates. Ablation studies further demonstrate that the attention mechanism is critical for effectively minimizing missed detection errors, and analysis of the learned attention weights reveals dataset-dependent layer importance patterns.

Future work includes extending LFA to foundation model backbones and to other safety-critical perception tasks, such as 3D object detection and multi-sensor fusion.

\bibliographystyle{IEEEbib}
\bibliography{strings,refs}

\end{document}